\documentclass[runningheads]{llncs}
\usepackage{graphicx}
\usepackage[dvipsnames]{xcolor}
\usepackage{subcaption}
\usepackage{booktabs}
\usepackage{multirow}
\usepackage[export]{adjustbox}

\begin{document}
\title{Happy Together: Learning and Understanding Appraisal From Natural Language}
\titlerunning{Learning and Understanding Appraisal From Natural Language}
%
\author{Arun Rajendran \and
Chiyu Zhang \and
Muhammad Abdul-Mageed}
\authorrunning{Rajendran et al.}
\institute{Natural Language Processing Lab\\
The University of British Columbia\\
\email{muhammad.mageed@ubc.ca}}
\maketitle              
\begin{abstract}

In this paper, we explore various approaches for learning two types of appraisal components from happy language. We focus on `agency' of the author and the `sociality' involved in happy moments based on the HappyDB dataset. We develop models based on deep neural networks for the task, including uni- and bi-directional long short-term memory networks, with and without attention. We also experiment with a number of novel embedding methods, such as embedding from neural machine translation (as in CoVe) and embedding from language models (as in ELMo). We compare our results to those acquired by several traditional machine learning methods. Our best models achieve 87.97\% accuracy on agency and 93.13\% accuracy on sociality, both of which are significantly higher than our baselines.


\end{abstract}
\section{Introduction}
Emotion is an essential part of human experience that affects both individual and group decision making. For this reason, it is desirable to understand the language of emotion and develop tools to aid such an understanding. Although there has been recently works focusing on detecting human emotion from text data ~\cite{volkova2016inferring,abdul2017emonet,alhuzali2018enabling}, we still lack a deeper understanding of various components related to emotion. Available emotion detection tools have so far been based on theories of basic emotion like the work of Paul Ekman and colleagues (e.g., ~\cite{ekman1992argument}) and extensions of these (e.g., Robert Plutchik's models ~\cite{plutchik1994psychology}). Emotion theory, however, has more to offer than mere categorization of human experience based on valence (e.g., anger, joy, sadness). As such, computational treatment of emotion is yet to benefit from existing (e.g., psychological) theories by building models that capture nuances these theories offer. Our work focuses on the cognitive appraisal theory ~\cite{roseman1984cognitive} where Roseman posits the existence of 5 appraisal components, including that of `agency'. Agency refers to whether a stimuli is caused by the individual, \textit{self-caused}, another individual, \textit{other-caused}, or merely the result of the situation \textit{circumstance-caused}. Identifying the exact type of agency related to an emotion is useful in that it helps determine the target of emotion (i.e., another person or some other type of entity). We focus on agency since it was recently labeled as an extension of the HappyDB dataset ~\cite{asai2018happydb} as part of the CL-Aff shared task~\cite{overview_claff}~\footnote{\url{https://sites.google.com/view/affcon2019/cl-aff-shared-task}}. 

The CL-Aff shared task distribution of HappyDB also includes labels for the concept of `sociality'. Sociality refers to whether or not other people than the author are involved in the emotion situation. Identifying the type of sociality associated to an emotion further enriches our knowledge of the emotion experience. For example, an emotion experience with a sociality value ``yes" (i.e., other people are involved) could teach us about social groups (e.g., families) and the range of emotions expressed during specific types of situations (e.g., wedding, death). Overall, agency and sociality are two concepts that we believe to be useful. Predictions of these concepts can be added to a computational toolkit which can be run on huge datasets to derive useful insights. To the best of our knowledge, no works have investigated learning these two concepts from language data. In this paper, we thus aim at pioneering this learning task by developing novel deep learning models for predicting agency and sociality. 

Moreover, we train attention-based models that are able to assign weights to features contributing to a given task. In other words, we are able to identify the words most relevant to each of the two concepts of agency and sociality. This not only enriches our knowledge about the distribution of these language items over each of these concepts, but also provides us with intuition about what our models learn (i.e., model interpretability). Interpretability is becoming increasingly important for especially deep learning models since many of these models are currently deployed in various real-life domains. Being able to identify why a model is making a certain decision helps us explain model decisions to end users, including by showing them examples of attention-based outputs.

In modeling agency and sociality, we experiment with various machine learning methods, both traditional and deep learning-based. In this way, we are able to establish strong baselines for this task as well as report competitive models. Our deep learning models are based on recurrent neural networks ~\cite{graves2012supervised,hochreiter1997long}. We also exploit frameworks with novel embedding methods, including embeddings from neural machine translation as in CoVe ~\cite{mccann2017learned} and embedding from language models as in ELMo ~\cite{peters2018deep}. Additionally, we investigate the utility of fine-tuning our models using the recently proposed ULMFiT model ~\cite{howard2018universal}.

Overall, we offer the following contributions: (1) we develop successful models for identifying the novel concepts of agency and sociality in happy language, (2) we probe our models to offer meaningful interpretations (in the form of visualization) of the contribution of different words to the learning tasks, thereby supporting model interpretability. The rest of the paper is organized as follows: Section ~\ref{sec:data} is about our dataset and data splits. In Section ~\ref{sec:methods}, we describe our methods and in Section ~\ref{sec:results} we provide our results. We offer model attention-based visualizations in Section ~\ref{sec:viz}, and we conclude in Section ~\ref{sec:conc}.

\section{Dataset}\label{sec:data}

HappyDB ~\cite{asai2018happydb} is a dataset of about 100,000 `happy moments' crowd-sourced via Amazon’s Mechanical Turk where each worker was asked to describe in a complete sentence ``what made them happy in the past 24 hours". Each user was asked to describe three such moments. In particular, we exploit the agency and sociality annotations provided on the dataset as part of the recent CL-Aff shared task \footnote{\url{https://sites.google.com/view/affcon2019/cl-aff-shared-task?authuser=0}}, associated with the AAAI-19 workshop of affective content analysis \footnote{\url{https://sites.google.com/view/affcon2019/}}. 

For this particular shared task, 10,560 moments are labelled for agency and sociality and were available as labeled training data.~\footnote{There were also 72,326 moments available as unlabeled training data, but we did not use these in our work.} Then, there were 17,215 moments used as test data. Test labels were not released and teams were expected to submit the predictions based on their systems on the test split. For our models, we split the labeled data into 80\% training set (8,448 moments) and 20\% development set (2112 moments). We train our models on train and tune parameters on dev. For our system runs, we submit labels from the models trained only on the 8,448 training data points.
The distribution of labeled data is as follows: agency (`yes'=7,796; `no'= 2,764), sociality (`yes'=5,625; `no'= 4,935).


\section{Methods}\label{sec:methods}

\subsection{Traditional Machine learning Models}
We develop multiple basic machine learning models, including Naive Bayes, Linear Support Vector Machine (LinSVM), and Logistic Regression (Log Reg). For each model, we have two settings: (a) we use a bag of words (BOW) approach (with n-gram values from 1 to 4) and (2) we combine the BOW with a TF-IDF transformation. These are strong baselines, due to our use of the combination of higher up n-grams (with n=4). We use the default parameters of \textit{Scikit-learn}\footnote{\url{https://scikit-learn.org/}} to train all the classical machine learning models. We also use an ensemble method that takes prediction labels from each of the classifiers and finds the majority among the different model classifications to decide the final prediction. We report results in terms of binary classification accuracy.

\subsection{Deep Learning}
We apply various models based on deep neural networks. All our deep learning models are based on variations of recurrent neural networks (RNNs), which have achieved remarkable performance on text classification tasks such as sentiment analysis and emotion detection \cite{tai2015improved,ren2016context,liu2016recurrent,abdul2017emonet,xu2016cached,samy2018context}. RNNs and its variations are able to capture sequential dependencies especially in time series data. One weakness of basic RNNs, however, lies in the gradient either vanishing or exploding, as the time gaps become larger. Long short term memory (LSTM) networks \cite{hochreiter1997long} were developed to address this limitation. We also use a bidirectional LSTM (BiLSTM). BiLSTM extends the unidirectional LSTM network by offering a second layer where the hidden to hidden states flow in opposite chronological order \cite{zhou2016attention}. 
Overall, our systems can be categorized as follows: (1) Systems tuning simple pre-trained embeddings; (2) Systems tuning embeddings from neural machine translation (NMT); (3) Systems tuning embeddings from language models (LM); and (4) Systems directly tuning language models (ULMFiT). 

\subsubsection{Exploiting Simple GloVe Embeddings} \label{GloVe}
For the embedding layer, we obtain the 300-dimensional embedding vector for tokens using GloVe's Common Crawl pre-trained model \cite{pennington2014glove}. GloVe embeddings are global vectors for word representation based on frequencies of pairs of co-occurring words. In this setting, we fix the embedding layer in our deep learning models at pre-trained GloVe embeddings. We apply four architectures (i.e. LSTM, LSTM with attention, BiLSTM, and BiLSTM with attention) to learn classification of agency and sociality respectively. For each models, we optimize the number of layers and the number of hidden unit within each layer to obtain the best performance. We experiment with layers from the set \{1, 2\}, and hidden units from the set \{128, 256, 512\}. Each of the setting was run with batch size 64 and dropout 0.75 for 20 epochs. 

\subsubsection{Embeddings from NMT} \label{CoVe}
Bryan et al. ~\cite{mccann2017learned} proposed CoVe, an approach for contextualized word embeddings directly from machine translation models. CoVe not only contains the word-level information from GloVe but also information learned with an LSTM in the context of the MT task. CoVe is trained on three different MT datasets, 2016 WMT multimodal dataset, 2016 IWSLT training set, and 2017 WMT news track training set. 
To train CoVe, we use an LSTM with attention. Our hyperparameter using CoVe are shown in Table ~\ref{global_hyp}.

\begin{table}[]
\centering
\caption{\label{global_hyp}Hyperparameters of CoVe, ELMo, and ULMFiT}
\begin{tabular}{lrrr}
\hline
\textbf{Hyperparameter}   & \multicolumn{1}{l}{\textbf{CoVe}} & \multicolumn{1}{l}{\textbf{ELMo}} & \multicolumn{1}{l}{\textbf{ULMFiT}} \\ \hline
\textbf{Embedding Size}   & 1,500                             & 1,024                             & 400                                 \\
\textbf{Hidden Nodes}     & 1,024                             & 512                               & 1,150                               \\
\textbf{Number of Layers} & 1                                 & 1                                 & 3                                   \\
\textbf{Dropout}          & 0.75                              & 0.5                               & 0.75                                \\
\textbf{Batch Size}       & 64                                & 64                                & 64                                  \\
\textbf{Epochs}           & 4                                & 8                                & 15                                  \\ \hline
\end{tabular}
\end{table}

\subsubsection{Embedding from LM}  \label{ELMo}
Peters et al. \cite{peters2018deep,Gardner2017AllenNLP} introduced ELMo, a model based on learning embeddings directly from language models. The pre-training with language models provides ELMo with both complex characteristics of words as well as the usage of these words across various linguistic contexts. ELMo is trained on 1 billion words benchmark dataset \cite{chelba2013one}, and these embeddings are employed as our input layer. More specifically, we extract the 3rd layer of the ELMo representation and experiment with it using an LSTM with attention network. \textbf{\textit{This is our best model, and the only model we submitted for the competition. We provide its hyperparameters in Table \ref{global_hyp}.}}


\subsubsection{Fine Tuning LM: ULMFiT}  \label{ULMFiT}
Transfer learning is extensively used in the field of computer vision for improving the ability of models to learn on new data. Inspired by this idea, Howard and Ruder \cite{howard2018universal} present ULMFiT\footnote{\url{http://nlp.fast.ai/}}, fine tunes a pre-trained language model (trained on the Wikitext-103 dataset). With ULMFiT, we use a forward language model. We use the same network architecture and hyperparameters (except dropout ratio and epochs) that Howard and Ruder used, as we report in Table \ref{global_hyp}.

\section{Results}\label{sec:results}

Table \ref{traditionML} shows the accuracy score of our traditional machine learning models on the validation set for both agency and sociality. As Table \ref{traditionML} shows, our linear models perform very well on the task, compared to our majority class baselines.  

\begin{table}[]
\centering
\caption{\label{traditionML} Performance of traditional machine learning models. Baseline is the majority class in our training split}
\begin{tabular}{@{}lcc@{}}
\toprule
\multicolumn{1}{c}{\textbf{Task}}     & \textbf{Agency ~} & \textbf{Sociality} \\ \midrule
\textbf{Baseline}              & 0.7382   & 0.5327        \\
\textbf{Naive Bayes (BOW)}      & 0.8073    & 0.8613              \\
\textbf{LinSVM (BOW)}          & 0.8281       & 0.8935            \\
\textbf{Log Reg (BOW)}           & 0.8215      & 0.8930           \\
\textbf{NB (BOW+TF-IDF)}           & 0.7462    & 0.8428           \\
\textbf{LinSVM (BOW+TF-IDF)}  & \textbf{0.8381} & \textbf{0.8963}                   \\
\textbf{Log Reg (BOW+TF-IDF)}   & 0.8320    & 0.8925              \\
\textbf{Ensemble}               & 0.8357     & 0.8930            \\ \bottomrule
\end{tabular}
\end{table}

Tables \ref{deepMLAgency} and \ref{deepMLSocial} show the results, with different model settings, in accuracy for the agency and sociality prediction tasks, respectively,  when we exploit simple GloVe embeddings with our LSTM-based 4 models. For each network, we highlight the highest accuracy.  The highest accuracy for agency task in Table \ref{deepMLAgency} is acquired with a 2-layered LSTM with attention (LSTM-A) with 512 units in each layer (accuracy = 0.8561). As \ref{deepMLSocial} shows, a 1-layered LSTM with 256 units acquires the best accuracy (0.9181) on the sociality task. These results suggest that sociality is an easier task than the agency task. One confounding factor is that the sociality training data is more balanced than the agency training data (with majority class at 0.5327 for sociality vs. 0.7382 for agency).
\begin{table}[]
\centering
\caption{\label{deepMLAgency}Results on the agency task with simple GloVe embeddings}
\begin{tabular}{@{}cccccc@{}}
\toprule
Layers ~ & Nodes ~ & LSTM ~   & BiLSTM ~ & LSTM-A ~& BiLSTM-A \\ \midrule
1      & 128   & \textbf{0.8527} & 0.8333 & 0.8362    & 0.8310       \\
       & 256   & 0.8518 & 0.8447 & 0.8433    & \textbf{0.8461}      \\
       & 512   & 0.8385 & 0.8229 & 0.8561    & 0.8433      \\ \cmidrule(r){1-6}
2      & 128   & 0.8475 & 0.8319 & 0.8504    & 0.8177      \\
       & 256   & 0.8376 & 0.7495 & 0.8513    & 0.8319      \\
       & 512   & 0.8182 & \textbf{0.8385} & \textbf{0.8561}    & 0.8272      \\ \bottomrule
\end{tabular}
\end{table}


\begin{table}[]
\centering

\caption{\label{deepMLSocial}Results on the sociality task with simple GloVe embeddings}
\begin{tabular}{@{}cccccc@{}}
\toprule
Layers ~ & Nodes ~ & LSTM ~   & BiLSTM ~ & LSTM-A ~& BiLSTM-A \\ \midrule
1      & 128   & 0.9025 & 0.9015 & 0.9115    & 0.9020       \\
       & 256   & \textbf{0.9181} & 0.8996 & 0.9067    & 0.9105      \\
       & 512   & 0.9124 & 0.8996 & 0.9048    & 0.9025      \\ \cmidrule(r){1-6}
2      & 128   & 0.9190  & \textbf{0.9176} & 0.9119    & 0.9134      \\
       & 256   & 0.9072 & 0.9129 & \textbf{0.9124}    & \textbf{0.9167}      \\
       & 512   & 0.9105 & 0.9058 & 0.9077    & 0.9152      \\ \bottomrule
\end{tabular}
\end{table}

Next, we present our results with the CoVe, ELMo, and ULMFiT trained models in Table \ref{embeddingperformance}. Table \ref{embeddingperformance} shows results in accuracy, AUC score, and F1 score(for positive class in binary classification) of our models on the validation set. From Tables \ref{traditionML}, \ref{deepMLSocial},  \ref{deepMLAgency} and \ref{embeddingperformance}, it can be observed that our CoVe, ELMo and ULMFiT models lead to (a) significant performance improvements compared to traditional machine learning models and (b) sizable improvements compared to deep learning models with simple GloVe embeddings. 

Among the systems with pre-trained embeddings (mentioned in Section \ref{CoVe}), ELMo performs better best. One nuance is that ELMo outperform the ULMFiT model that fine-tunes a language model rather than the embeddings. One probable explanation for this is the impact of attention that is used in the LSTM model with ELMo embedding which is crucial for this particular task and is not present in the ULMFiT model. We now turn to probing our models further by visualizing the attention weights for words in our data.


\begin{table}[!htbp]
\centering
\caption{\label{embeddingperformance}Performance of systems with embeddings from pre-trianed models}
\begin{tabular}{@{}lccc|ccc@{}}
\toprule
\multicolumn{1}{c}{\multirow{2}{*}{\textbf{Model}}} & \multicolumn{3}{c}{\textbf{Agency}}                  & \multicolumn{3}{c}{\textbf{Sociality}}                  \\ \cmidrule(l){2-7} 
\multicolumn{1}{c}{}                                & \textbf{Accuracy} & \textbf{AUC} & \textbf{F1 Score} & \textbf{Accuracy} & \textbf{AUC} & \textbf{F1 Score} \\ \midrule
\textbf{Baseline}                                      & 0.7382            & 0.5       & 0.8494  & 0.5327            & 0.5       & 0.6951                       \\
\textbf{LinSVM(BOW+TF-IDF)}                                      & 0.8381            & 0.7713       & 0.8926 & 0.8963            & 0.8951       & 0.9037                        \\
\textbf{CoVe (LSTM-A)}                                      & 0.8726            & 0.8081       & 0.7345 & 0.9181            & 0.9180       & 0.9127                        \\
\textbf{ELMo (LSTM-A)}                                      & \textbf{0.8797}            & \textbf{0.8444}       & \textbf{0.9185}  & \textbf{0.9313}            & \textbf{0.9309}       & \textbf{0.9342}                      \\
\textbf{ULMFiT (LSTM)}                              
 & 0.8660            & 0.8277       & 0.9095 & 0.9237            & 0.9235       & 0.9285                      \\
\bottomrule
\end{tabular}
\end{table}
\section{Attention-Based Visualization}\label{sec:viz}
For interpretability, and to acquire a better understanding of the two important concepts of agency and sociality, we provide attention-based visualization of 24 example sentences from our data. In each example, color intensity corresponds to the self attention weights assigned by our model (LSTM-A). Figures 1 (hand-picked) and 2 (randomly picked) provide examples from the agency data, each for the positive then the negative class respectively. As the Figures demonstrate, the model attentions are relatively intuitive. For example, for the positive class cases (hand-picked), the model attends to words such as `my', `with', and `co-workers', which refer to (or establishes a connection with) the agent. Figures 3 and 4 provide similar visualizations for the sociality task. Again, the attention weights cast some intuitive light on the concept of sociality. The model, for example, attends to words like `daughter', `grandson', `members', and `family' in the hand-picked positive cases. Also, in the hand-picked negative examples, the model attends to words referring to non-persons such as `book', `mail', `workout', and `dog'. 

\begin{figure*}[h!]
  \centering
    \begin{subfigure}[b]{0.9\linewidth}
    \includegraphics[width=\linewidth,frame]{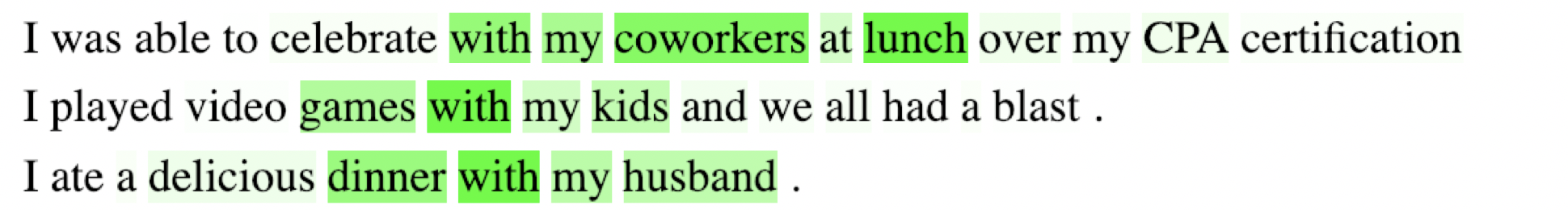}
    \caption{\textbf{\color{OliveGreen}{a. Examples of happy moments with \textit{positive} agency label}}}
  \end{subfigure}
    \begin{subfigure}[b]{0.9\linewidth}
    \includegraphics[width=\linewidth,frame]{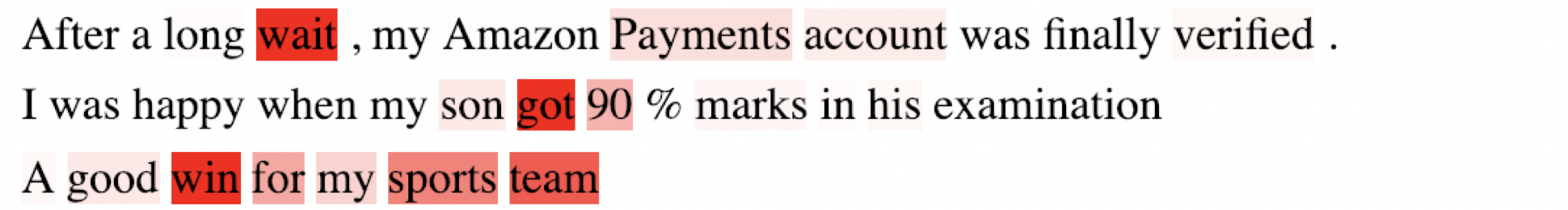}
    \caption{\textbf{\color{red}{b. Examples of happy moments with \textit{negative} agency label}}}
  \end{subfigure}
  \caption{ \setcounter{figure}{1} \textbf{Attention heatmap for hand-picked agency examples}}
  
\end{figure*}
\begin{figure*}[h!]
  \centering
    \begin{subfigure}[b]{0.9\linewidth}
    \includegraphics[width=\linewidth,frame]{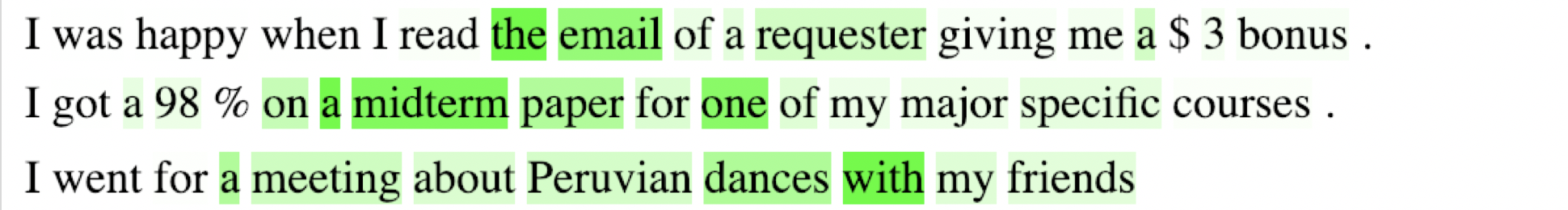}
    \caption{\textbf{\color{OliveGreen}{a. Examples of happy moments with \textit{positive} agency label}}}
  \end{subfigure}
    \begin{subfigure}[b]{0.9\linewidth}
    \includegraphics[width=\linewidth,frame]{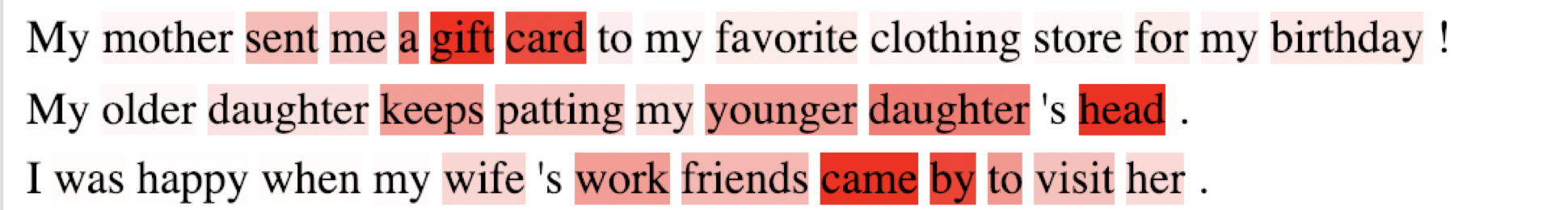}
    \caption{\textbf{\color{red}{b. Examples of happy moments with \textit{negative} agency label}}}
  \end{subfigure}
  \caption{ \setcounter{figure}{2} \textbf{Attention heatmap for randomly-picked agency examples}}
  \label{fig:agencyvis_random}
\end{figure*}
\begin{figure}[h!]
  \centering
    \begin{subfigure}[b]{0.9\linewidth}
        \includegraphics[width=\linewidth, frame]{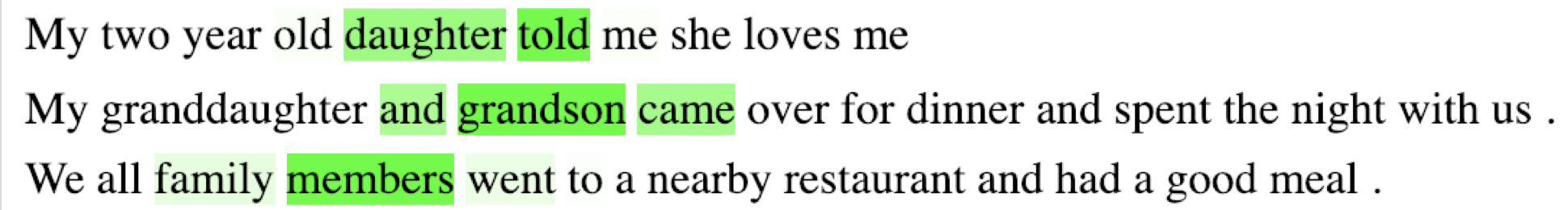}
        \subcaption{\textbf{\color{OliveGreen}{a. Examples of happy moments with \textit{positive} sociality label}}}
    \end{subfigure}
  
    \begin{subfigure}[b]{0.9\linewidth}
        \includegraphics[width=\linewidth,frame]{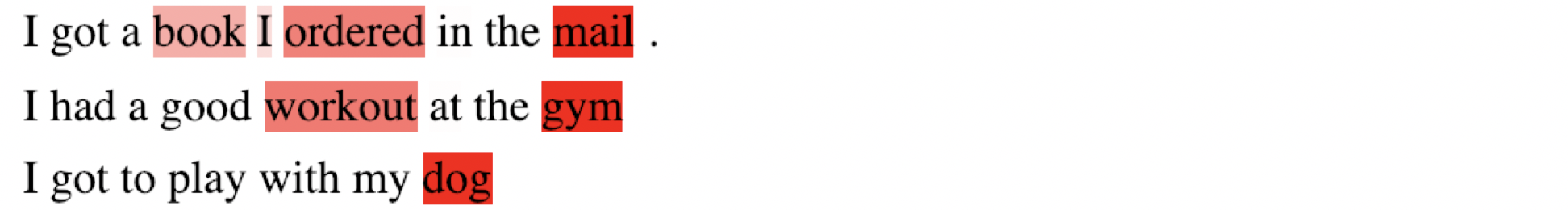}
        \subcaption{\textbf{\color{red}{b. Examples of happy moments with \textit{negative} sociality label}}}
  \end{subfigure}

  \caption{ \setcounter{figure}{3} \textbf{Attention heatmap for hand-picked sociality examples}}
  \label{fig:socialvis_hand}
\end{figure}
\begin{figure*}[h!]
  \centering
    \begin{subfigure}[b]{0.9\linewidth}
    \includegraphics[width=\linewidth, frame]{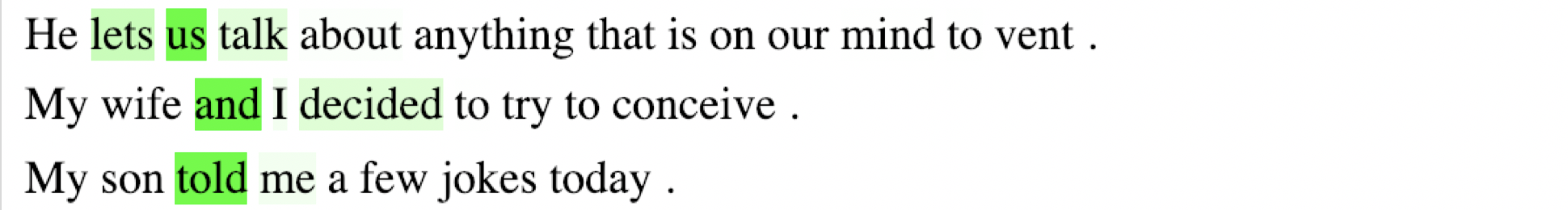}
    \caption{\textbf{\color{OliveGreen}{a. Examples of happy moments with \textit{positive} sociality label}}}
  \end{subfigure}
    \begin{subfigure}[b]{0.9\linewidth}
    \includegraphics[width=\linewidth,frame]{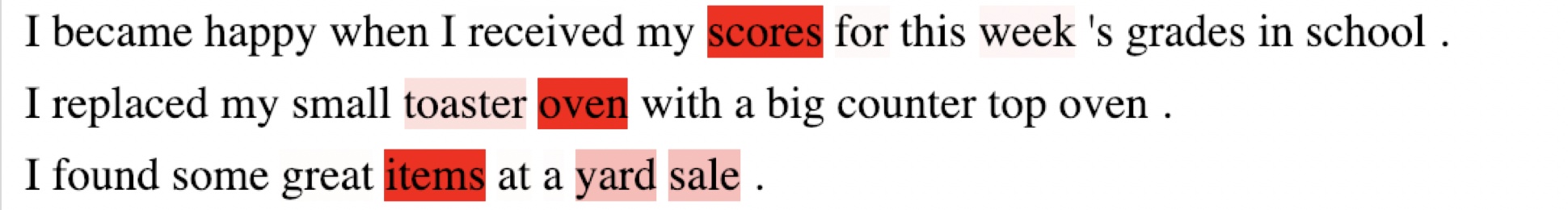}
    \caption{\textbf{\color{red}{b. Examples of happy moments with \textit{negative} sociality label}}}
  \end{subfigure}
  \caption{ \setcounter{figure}{4} \textbf{Attention heatmap for randomly-picked sociality examples}}
  \label{fig:socialvis_random}
\end{figure*}

\begin{figure*}
\centering
  \includegraphics[width=\linewidth]{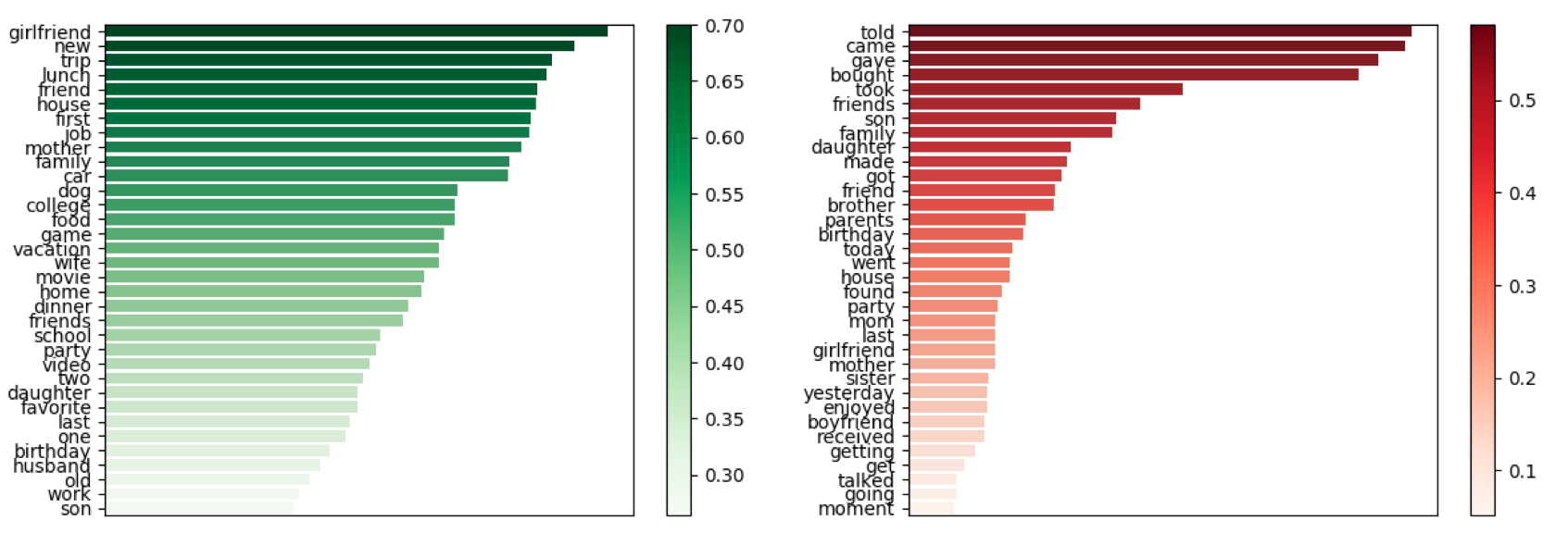}
  \captionsetup{justification=centering}
  \caption{Importance ranking of words in happy moments with \textit{positive} agency (\textbf{left}) and \textit{positive} sociality (\textbf{right})}.
  \label{fig:rank_social_agency}
\end{figure*}

Next, in Figure ~\ref{fig:rank_social_agency}, we provide the top 35 words the model attends to in the positive classes in each of the agency and sociality datasets. Again, many of these words are intutively relevant to each task. For example, for agency, the model attends to words referring to others the agent indicating interaction with (e.g., `girlfriend', `friend' `mother' and `family') and social activities the agent is possibly contributing to (e.g., `lunch', 'trip', `party', and `dinner'). Similarly, for sociality, the model is attending to verbs indicating being socially involved (e.g., `told', `came', `bought', and `took') and others/social groups (e.g., `friends', `son', `family', and `daughter'). Clearly, the two concepts of agency and sociality are not orthogonal: The words the model attends to in each case indicate overlap between the two concepts to some extent.

\section{Conclusion}\label{sec:conc}
In this paper, we reported successful models learning agency and sociality in a supervised setting. We also presented extensive visualizations based on the models' self-attentions that enhance our understanding of these two concepts as well as model decisions (i.e., interpretability). In the future, we plan to develop models for the same tasks based on more sophisticated attention mechanisms.

\section{Acknowledgement}
We acknowledge the support of the Natural Sciences and Engineering Research Council of Canada (NSERC). The research was partially enabled by support from WestGrid (\url{www.westgrid.ca}) and Compute Canada (\url{www.computecanada.ca}).

\bibliographystyle{splncs04}
\bibliography{llncs.bib}

\begin{thebibliography}{10}
\providecommand{\url}[1]{\texttt{#1}}
\providecommand{\urlprefix}{URL }
\providecommand{\doi}[1]{https://doi.org/#1}

\bibitem{abdul2017emonet}
Abdul-Mageed, M., Ungar, L.: Emonet: Fine-grained emotion detection with gated
  recurrent neural networks. In: Proceedings of the 55th Annual Meeting of the
  Association for Computational Linguistics (Volume 1: Long Papers). vol.~1,
  pp. 718--728 (2017)

\bibitem{alhuzali2018enabling}
Alhuzali, H., Abdul-Mageed, M., Ungar, L.: Enabling deep learning of emotion
  with first-person seed expressions. In: Proceedings of the Second Workshop on
  Computational Modeling of People’s Opinions, Personality, and Emotions in
  Social Media. pp. 25--35 (2018)

\bibitem{asai2018happydb}
Asai, A., Evensen, S., Golshan, B., Halevy, A., Li, V., Lopatenko, A.,
  Stepanov, D., Suhara, Y., Tan, W.C., Xu, Y.: Happydb: A corpus of 100,000
  crowdsourced happy moments. arXiv preprint arXiv:1801.07746  (2018)

\bibitem{chelba2013one}
Chelba, C., Mikolov, T., Schuster, M., Ge, Q., Brants, T., Koehn, P., Robinson,
  T.: One billion word benchmark for measuring progress in statistical language
  modeling. arXiv preprint arXiv:1312.3005  (2013)

\bibitem{ekman1992argument}
Ekman, P.: An argument for basic emotions. Cognition \& emotion
  \textbf{6}(3-4),  169--200 (1992)

\bibitem{Gardner2017AllenNLP}
Gardner, M., Grus, J., Neumann, M., Tafjord, O., Dasigi, P., Liu, N.F., Peters,
  M., Schmitz, M., Zettlemoyer, L.S.: {AllenNLP}: A deep semantic natural
  language processing platform. In: ACL workshop for NLP Open Source Software
  (2018)

\bibitem{graves2012supervised}
Graves, A.: Supervised sequence labelling. In: Supervised Sequence Labelling
  with Recurrent Neural Networks, pp. 5--13. Springer (2012)

\bibitem{hochreiter1997long}
Hochreiter, S., Schmidhuber, J.: Long short-term memory. Neural computation
  \textbf{9}(8),  1735--1780 (1997)

\bibitem{howard2018universal}
Howard, J., Ruder, S.: Universal language model fine-tuning for text
  classification. In: Proceedings of the 56th Annual Meeting of the Association
  for Computational Linguistics (Volume 1: Long Papers). vol.~1, pp. 328--339
  (2018)

\bibitem{overview_claff}
Jaidka, K., Mumick, S., Chhaya, N., Ungar, L.: {The CL-Aff Happiness Shared
  Task: Results and Key Insights}. In: Proceedings of the 2nd Workshop on
  Affective Content Analysis @ AAAI {(AffCon2019)}. Honolulu, Hawaii (January
  2019)

\bibitem{liu2016recurrent}
Liu, P., Qiu, X., Huang, X.: Recurrent neural network for text classification
  with multi-task learning. In: Proceedings of the Twenty-Fifth International
  Joint Conference on Artificial Intelligence. pp. 2873--2879. AAAI Press
  (2016)

\bibitem{mccann2017learned}
McCann, B., Bradbury, J., Xiong, C., Socher, R.: Learned in translation:
  Contextualized word vectors. In: Advances in Neural Information Processing
  Systems. pp. 6294--6305 (2017)

\bibitem{pennington2014glove}
Pennington, J., Socher, R., Manning, C.D.: Glove: Global vectors for word
  representation. In: EMNLP. vol.~14, pp. 1532--1543 (2014)

\bibitem{peters2018deep}
Peters, M.E., Neumann, M., Iyyer, M., Gardner, M., Clark, C., Lee, K.,
  Zettlemoyer, L.: Deep contextualized word representations. arXiv preprint
  arXiv:1802.05365  (2018)

\bibitem{plutchik1994psychology}
Plutchik, R.: The psychology and biology of emotion. HarperCollins College
  Publishers (1994)

\bibitem{ren2016context}
Ren, Y., Zhang, Y., Zhang, M., Ji, D.: Context-sensitive twitter sentiment
  classification using neural network. In: AAAI. pp. 215--221 (2016)

\bibitem{roseman1984cognitive}
Roseman, I.J.: Cognitive determinants of emotion: A structural theory. Review
  of personality \& social psychology  (1984)

\bibitem{samy2018context}
Samy, A.E., El-Beltagy, S.R., Hassanien, E.: A context integrated model for
  multi-label emotion detection. Procedia computer science  \textbf{142},
  61--71 (2018)

\bibitem{tai2015improved}
Tai, K.S., Socher, R., Manning, C.D.: Improved semantic representations from
  tree-structured long short-term memory networks. arXiv preprint
  arXiv:1503.00075  (2015)

\bibitem{volkova2016inferring}
Volkova, S., Bachrach, Y.: Inferring perceived demographics from user emotional
  tone and user-environment emotional contrast. In: Proceedings of the 54th
  ACL. vol.~1, pp. 1567--1578 (2016)

\bibitem{xu2016cached}
Xu, J., Chen, D., Qiu, X., Huang, X.: Cached long short-term memory neural
  networks for document-level sentiment classification. In: Proceedings of the
  2016 Conference on Empirical Methods in Natural Language Processing. pp.
  1660--1669 (2016)

\bibitem{zhou2016attention}
Zhou, P., Shi, W., Tian, J., Qi, Z., Li, B., Hao, H., Xu, B.: Attention-based
  bidirectional long short-term memory networks for relation classification.
  In: Proceedings of the 54th Annual Meeting of the Association for
  Computational Linguistics (Volume 2: Short Papers). vol.~2, pp. 207--212
  (2016)

\end{thebibliography}
\end{document}